\documentclass[a4paper]{article}
\usepackage{graphicx,spconf,amsmath,amssymb}
\usepackage{subfigure,algorithm,algorithmic}
\usepackage{xcolor}
\usepackage{flushend}

\usepackage{fancyheadings}
\title{\Large Barcodes for Medical Image Retrieval \\ Using Autoencoded Radon Transform }

\name{Hamid R. Tizhoosh$^1$, Christopher Mitcheltree$^2$, Shujin Zhu$^{3,4}$, Shamak Dutta$^2$} 
\address{$^1$ KIMIA Lab, University of Waterloo, Canada\\
$^2$, Electrical and Computer Engineering, University of Waterloo, Canada \\
$^3$ School of Electronic \& Optical Eng., Nanjing University of Science \& Technology, Jiangsu, China\\
$^4$ Systems Design Engineering, University of Waterloo, Canada}

\begin{document}
\maketitle
\begin{abstract}
Using content-based binary codes to tag digital images has emerged as a promising retrieval technology. Recently, Radon barcodes (RBCs) have been introduced as a new binary descriptor for image search. RBCs are generated by binarization of Radon projections and by assembling them into a vector, namely the barcode. A simple local thresholding has been suggested for binarization. In this paper, we put forward the idea of ``autoencoded Radon barcodes''. Using images in a training dataset, we autoencode Radon projections to perform binarization on outputs of hidden layers. We employed the mini-batch stochastic gradient descent approach for the training. Each hidden layer of the autoencoder can produce a barcode using a threshold determined based on the range of the logistic function used. The compressing capability of autoencoders apparently reduces the redundancies inherent in Radon projections leading to more accurate retrieval results. The IRMA dataset with 14,410 x-ray images is used to validate the performance of the proposed method. The experimental results, containing comparison with RBCs, SURF and BRISK, show that autoencoded Radon barcode (ARBC) has the capacity to  capture important information and to learn richer representations resulting in lower retrieval errors for image retrieval measured with the accuracy of the first hit only.
\end{abstract}


\section{Introduction}
In recent years, the number of digital images in private and public archives has been steadily growing due to  advances in imaging and communication technologies. When a query image is given, the purpose of ``content-based image retrieval'' (CBIR) is to find the most similar images in the database. Here, the general assumption is that images are not annotated with descriptive texts and keywords, or it is more desirable to search within the image due to domain requirements, e.g., in medical imaging. CBIR systems, hence, need to numerically characterize the image content somehow, i.e., through some type of feature extractions, in order to facilitate the search queries. More recently, CBIR methods have shifted toward ``binary'' descriptors. Binary information is compact, and its processing is fast. One of the most recent proposals is to use Radon barcodes. Based on a well-established transform with diverse applications, among others in medical imaging, one can assemble binary vectors by proper thresholding of Radon projections. 

In this paper, we propose to train ``autoencoders'' to generate Radon barcodes. Presently, simple local thresholding is used to binarize Radon projections. This, as we will demonstrate, causes loss of information and hence an increase in retrieval error. By employing autoencoders we binarize the compressed version of Radon projections with less loss resulting in higher retrieval accuracy. As the Radon projections of neighbouring angles are highly redundant, the proposed approach constitutes a neural approach to redundancy reduction.   
           
In the following sections, we first briefly review the relevant literature in Section \ref{sec:BGK}. The proposed autoencoded Radon barcodes are described in Section \ref{sec:prop}. The dataset, error measurements, settings of the autoencoders, and the experimental results are reported in Section \ref{sec:exp}.

\section{Background Review}
\label{sec:BGK}
In this section, we briefly review the relevant literature on image retrieval, autoencoders, and Radon barcodes. 

\textbf{Image Retrieval --} Recent annotation or tagging methods for image retrieval have moved toward binary descriptors. Binary codes require less space and can be processed fast. Calonder et al. used binary strings and developed an efficient feature point descriptor called BRIEF  \cite{calonder2010brief}. Rublee et al. proposed another binary descriptor called ORB based on BRIEF, which should be rotation invariant as well as resistant to noise \cite{rublee2011orb}. Leutenegger et al. proposed a novel method, named Binary Robust Invariant Scalable Keypoints (BRISK) for image key point detection, description and matching \cite{leutenegger2011brisk}, and reported good performance at a low computational cost. In 2015, Radon barcodes based on the Radon transform was introduced \cite{tizhoosh2015barcode}; good overall results for medical image retrieval on the IRMA dataset were reported. Recently, autoencoders have been increasingly used for image retrieval tasks \cite{hinton2006reducing,camlica2015autoencoding}. Krizhevsky et al. first applied a deep convolutional neural network to the ImageNet LSVRC-2010 dataset and achieved the best result in that challenge \cite{krizhevsky2012imagenet}.

\textbf{Autoencoders --} The autoencoder is a type of artificial neural network which is trained to encode the input into some representations such that the input can be reconstructed from that representation \cite{bengio2009learning}. An autoencoder transforms an input \textbf{x} into hidden representation \textbf{y} through a deterministic mapping
\begin{equation}
\textbf{y}= s\left(\textbf{W}\textbf{x}+\textbf{b}\right),
\end{equation}
where $s$ is a non-linear function (e.g., Sigmoid), $\textbf{W}$ is a $d'\!\times\! d$ weight matrix, and $\textbf{b}$ is an offset vector of length $d'$. The prediction of original input can be reconstructed by mapping back the hidden representation:
\begin{equation}
\textbf{z}= s\left(\textbf{W}'\textbf{y}+\textbf{b}'\right).
\end{equation}

Generally, the reconstruction $\textbf{z}$ is not to be interpreted as exactly the same as the input $\textbf{x}$, ``but rather in probabilistic terms as the parameters (typically the mean) of a distribution that may generate \textbf{x} with high probability'' \cite{vincent2010stacked}.  In order to minimize the reconstruction error, the \textit{squared errors} $L(\textbf{x},\textbf{z})$ can be used:
\begin{equation}
L(\textbf{x},\textbf{z})=\left\|\textbf{x}-\textbf{z}\right\|^2.
\end{equation}
%
%\begin{equation}
%L_H(\textbf{x},\textbf{z})=-\sum^{d}_{k=1}\left[\textbf{x}_k log \textbf{z}_k+(1-\textbf{x}_k)log(1-\textbf{z}_k)\right]
%\end{equation}

\begin{figure}[htb]
\begin{center}
\vspace{0.05in}
\includegraphics[width=0.9\columnwidth]{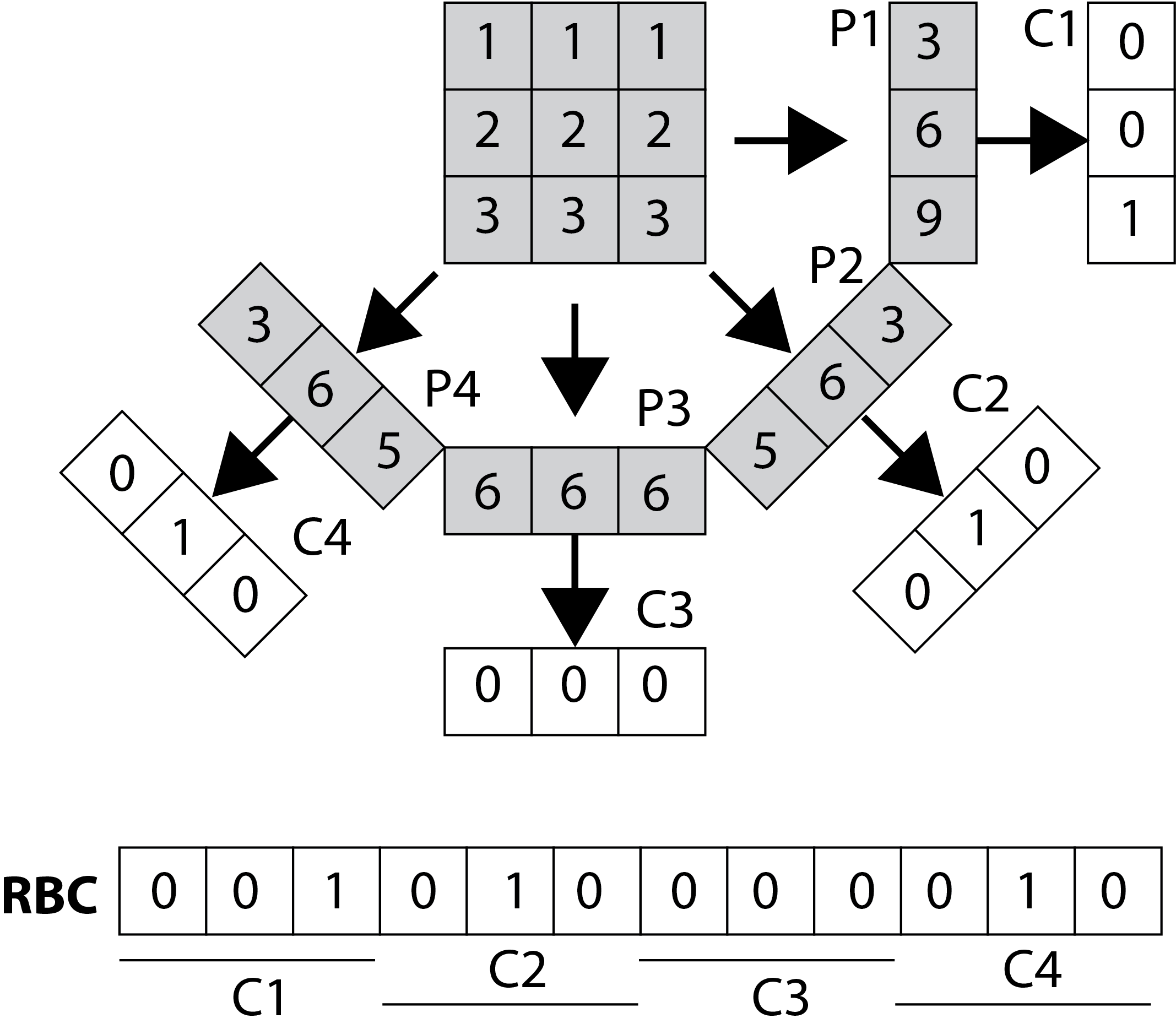}
\caption{All projections (P1,P2,P3,P4) generated by the Radon transform are thresholded to generate code fragments C1,C2,C3,C4 resulting in a barcode [C1 C2 C3 C4] [Source: \cite{tizhoosh2015barcode}].}
\label{fig:RBC}
\end{center}
\end{figure}

\textbf{Radon Barcodes --} The idea behind Radon barcodes (RBC) \cite{tizhoosh2015barcode,TizhooshISBI} is to apply Radon transform on an image and to binarize the projections in a proper way with minimum information loss. If the image $I$ is regarded as a 2D function $f(x,y)$, then one can project $f(x,y)$ along a number of projection angles $\theta$. The projection is basically the sum (integral) of $f(x,y)$ values along lines constituted by each angle $\theta$. The projection creates a new image $R(\rho,\theta)$ with $\rho = x \cos \theta + y \sin \theta$. Hence, using the Dirac delta function $\delta(\cdot)$ the Radon transform can be written as 
\begin{equation}
R(\rho,\theta) = \int\limits_{-\infty}^{+\infty} \int\limits_{-\infty}^{+\infty} f(x,y) \delta(\rho-x\cos \theta-y\sin\theta) dx dy.
\end{equation}
It has been proposed to threshold all projections for each angle based on a ``local'' threshold for that angle, such that a barcode of all thresholded projections can be generated (Figure \ref{fig:RBC}). Further, it has been proposed to use a typical value (e.g.,  median value of all non-zero values of each projection) as the threshold \cite{tizhoosh2015barcode}. Algorithm \ref{alg:Radon} describes how \textbf{Radon barcodes (RBC)} are generated.  In order to receive same-length barcodes \emph{Normalize$(I)$} resizes all images into $R_N\!\times\!C_N$ images (i.e., $R_N\!=\!C_N\!=\!2^n,n\in \mathbb{N}^+$).

% Radon Barcode -------------------------
\begin{algorithm}[htb]
\caption{Radon Barcodes via thresholding \cite{tizhoosh2015barcode}}
\begin{algorithmic}[1]
\label{alg:Radon}
\STATE Initialize Radon Barcode $\mathbf{r} \leftarrow \emptyset$ 
\STATE Initialize angle $\theta \leftarrow 0$ and $R_N=C_N\leftarrow 32$
\STATE Normalize the input image $\bar{I} = \textrm{Normalize}(I,R_N,C_N)$ 
\STATE Set the number of projection angles, e.g. $n_p \leftarrow 8$
\WHILE{$\theta < 180$}
	\STATE Get all projections $\mathbf{p}$ for $\theta$
	\STATE Find typical value $T_\textrm{typical}\leftarrow\textrm{median}_i (\mathbf{p}_i)|_{\mathbf{p}_i \neq 0}$
	\STATE Binarize projections: $\mathbf{b} \leftarrow \mathbf{p} \geq T_\textrm{typical}$ 
	\STATE Append the new row $\mathbf{r} \leftarrow \textrm{append}(\mathbf{r},\mathbf{b} )$ 
	\STATE $\theta \leftarrow \theta + \frac{180}{n_p}$
\ENDWHILE
\STATE Return  $\mathbf{r}$
 \end{algorithmic}
 \end{algorithm}

 The main difference between this work (described in the following section) and the original work on Radon barcodes  in \cite{tizhoosh2015barcode} is that we use the compressing capability of autoencoders in order to reduce the redundancies inherent in Radon projections. This clearly leads to, as we will report in the experiment section, more accurate retrieval results. 
 
\section{The Proposed Method}
\label{sec:prop}
We first pre-process images to subsequently autoencode them with a suitable architecture and setting of the autoencoder.

\textbf{Pre-Processing of Images --} In order to generate same-length inputs for the autoencoder, all  images in the training dataset are resized to fixed dimensions (e.g., $32\!\times\!32$). Radon transform with selected number of projection angles $n_\theta$ is employed to extract the Radon features (i.e., the projections). This representation of each image is then normalized to only consist of values between 0 and 1 by dividing each element in the feature vector by the maximum value of that projection angle. The normalized Radon features can then be fed to the autoencoder. 

\textbf{Autoencoded Radon Barcodes --} After the autoencoder is trained, the barcode can be generated by feeding each normalized Radon feature vector to the autoencoder and looking at the output of each hidden layer (the value of the Sigmoid function of each neuron in that layer) that is passed on to the next layer. If this value is greater than 0.5, it becomes a ``1'' in the barcode and if it is less than 0.5, it becomes a ``0''. The barcode can be generated for every hidden layer in the autoencoder.  

\begin {figure*}[htb]
\centering
\vspace{0.05in}
{\includegraphics[width=0.9\textwidth]{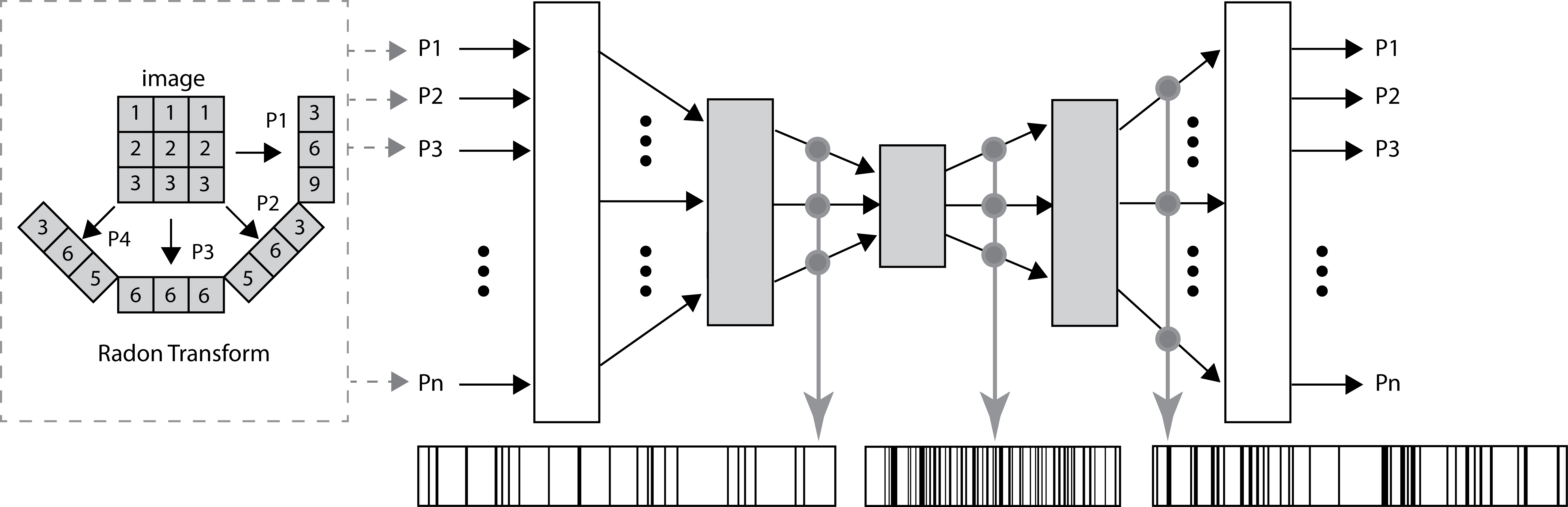}}\quad
\caption{Schematic illustration of the proposed method. The Radon projections ($P_1,P_2,\dots$) of the preprocessed image are autoencoded. For $n$ hidden layers, $n$ barcodes can be generated.}
\label{fig_flowchart}
\end{figure*}

\begin{figure*}[htb]
\centering
\subfigure{\includegraphics[width=3.8cm,height=3.8cm]{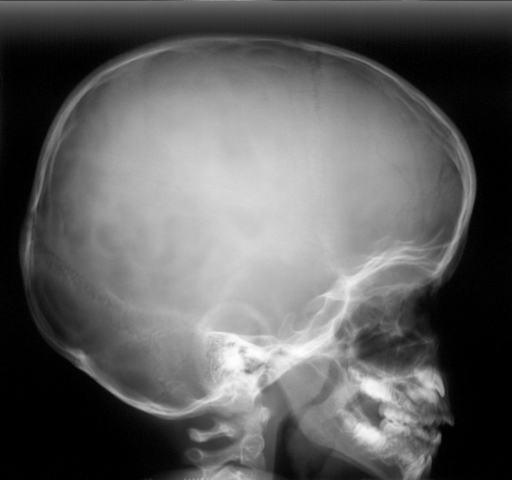}}\hspace{0.2cm}
\subfigure{\includegraphics[width=3.8cm,height=3.8cm]{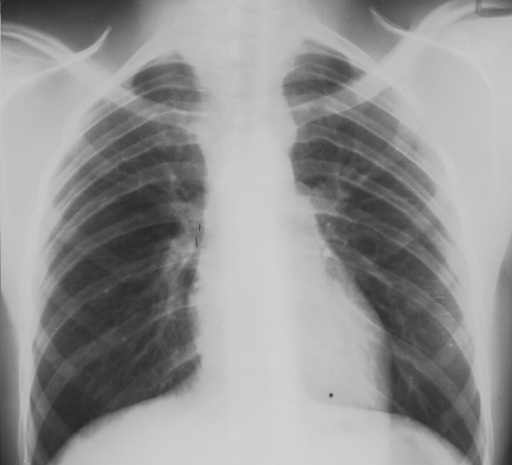}}\hspace{0.2cm}
\subfigure{\includegraphics[width=3.8cm,height=3.8cm]{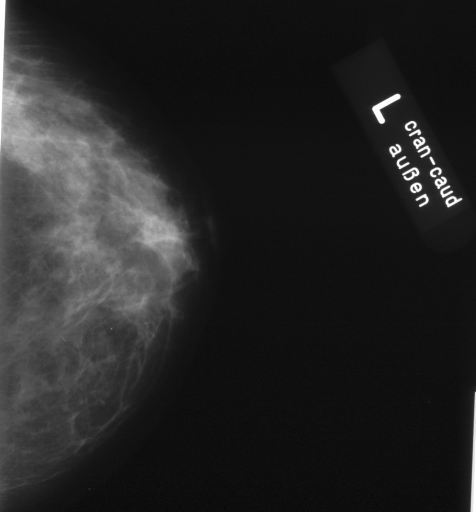}}\hspace{0.2cm}
\subfigure{\includegraphics[width=3.8cm,height=3.8cm]{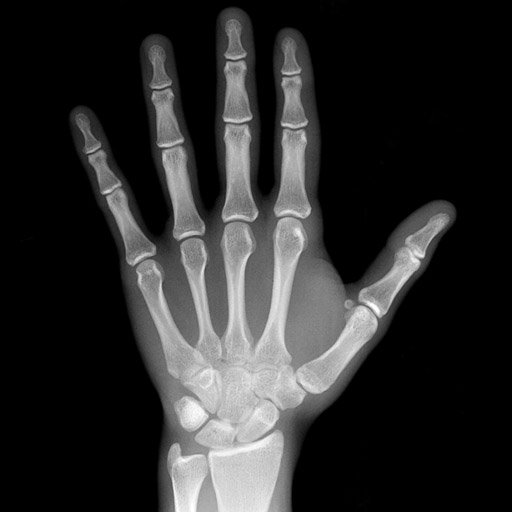}}\hspace{0.2cm}\quad

\subfigure{\includegraphics[width=3.8cm,height=1.0cm]{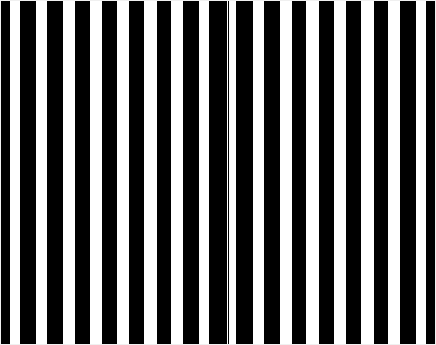}}\hspace{0.2cm}
\subfigure{\includegraphics[width=3.8cm,height=1.0cm]{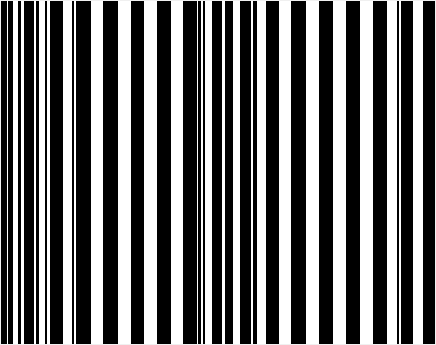}}\hspace{0.2cm}
\subfigure{\includegraphics[width=3.8cm,height=1.0cm]{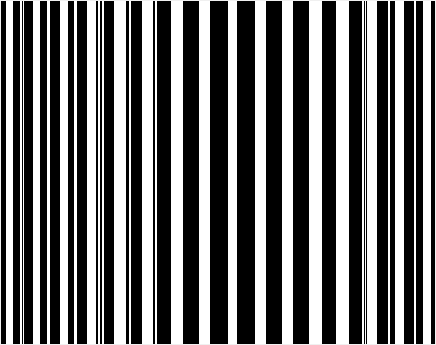}}\hspace{0.2cm}
\subfigure{\includegraphics[width=3.8cm,height=1.0cm]{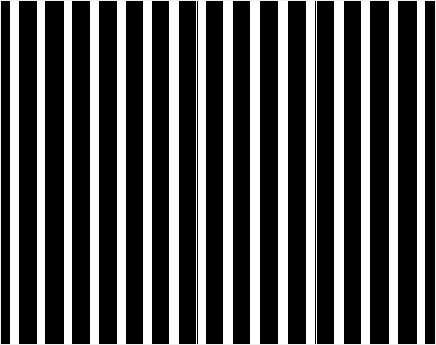}}\hspace{0.2cm}\quad

\subfigure{\includegraphics[width=3.8cm,height=1.0cm]{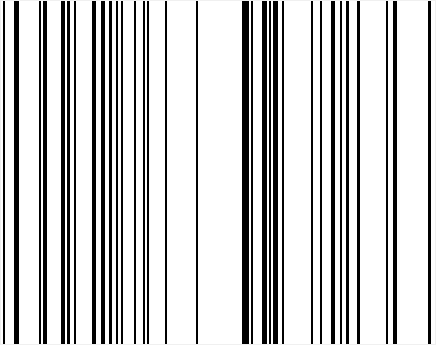}}\hspace{0.2cm}
\subfigure{\includegraphics[width=3.8cm,height=1.0cm]{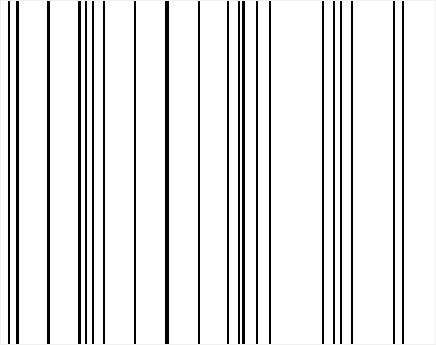}}\hspace{0.2cm}
\subfigure{\includegraphics[width=3.8cm,height=1.0cm]{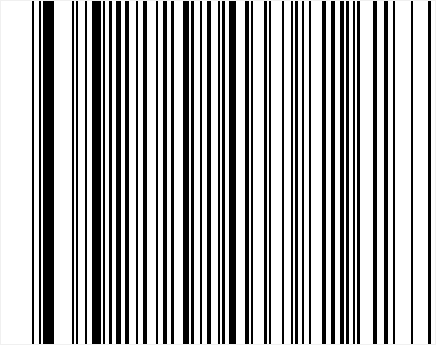}}\hspace{0.2cm}
\subfigure{\includegraphics[width=3.8cm,height=1.0cm]{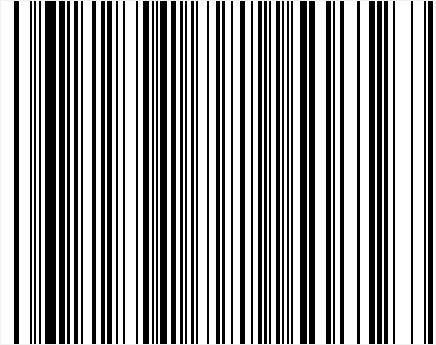}}\hspace{0.2cm}\quad
\caption{Visual comparison of Radon barcode (top) and autoencoder Radon barcode (bottom) for  sample x-ray images from IRMA dataset. Images are normalized to $32\times 32$ with 16 projections.}
\label{fig_rbc_arbc}
\end{figure*}

Figure \ref{fig_flowchart} illustrates the proposed method. Figure \ref{fig_rbc_arbc} shows four sample images from IRMA dataset with their Radon barcodes (with local thresholding) and the autoencoded Radon barcodes. For a better display, both barcodes are displayed with the same length (the length of autoencoded Radon barcodes are $1/4$ of the length of Radon barcodes).

\textbf{Autoencoder Setting --} We use the traditional autoencoder instead of the optimized one which trains fairly slowly and does not provide the most accurate results when compared to a much more advanced autoencoder \footnote{We used a Python implementation (unoptimized) that can be found here: https://goo.gl/lFdwVn}. The standard learning algorithm for neural networks, known as mini-batch stochastic gradient descent \cite{Li2014} is adopted. No regularization techniques were implemented. The weights of the autoencoder were initialized using a Gaussian distribution, $\mathcal{N}(0,1)$,  over the square root of the number of weights connecting to the same neuron and the biases were initialized using a Gaussian distribution with mean 0 and standard deviation 1. We use the mean squared error as the cost function. Other parameters such as learning rate, mini-batch size and training epochs are set empirically as described in the next section.

\section{Experimental Results}
\label{sec:exp}
We first describe the IRMA dataset that we used for experimentation. The error measurement will be describe next. Subsequently, we report the parameter settings for the autoencoder. Finally, we analyze the performance of the proposed approach based on the experimental results. A comparison with SURF and BRISK methods will be reported at the end. 

\begin{figure*}[htb]
\centering     %%% not \center
\vspace{0.05in}
\subfigure[1121-127-700-500]{\label{fig:a}\includegraphics[width=35mm,height=35mm]{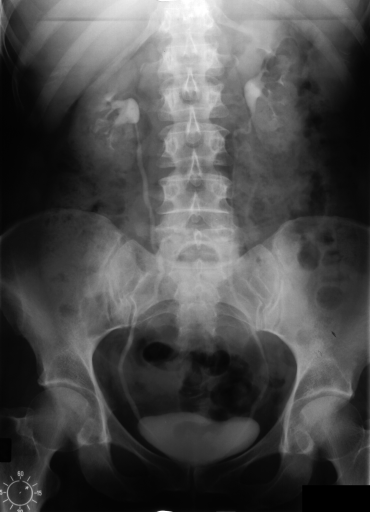}} \quad
\subfigure[1121-120-942-700]{\label{fig:b}\includegraphics[width=35mm,height=35mm]{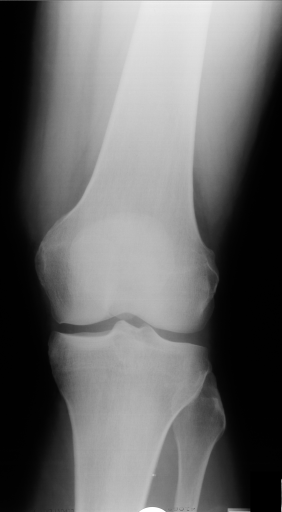}} \quad
\subfigure[1121-120-200-700]{\label{fig:b}\includegraphics[width=35mm,height=35mm]{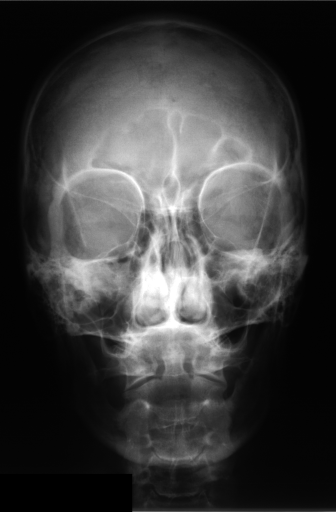}} \quad
\subfigure[1121-120-918-700]{\label{fig:b}\includegraphics[width=35mm,height=35mm]{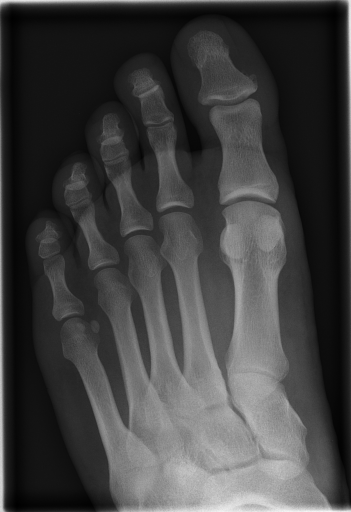}}\\
\caption{Sample images from IRMA Dataset with their IRMA codes TTTT-DDD-AAA-BBB.}
\label{fig:IRMASamples}
\end{figure*}

\textbf{Image Test Data --} The Image Retrieval in Medical Applications (IRMA) database\footnote{http://irma-project.org/} is a collection of more than 14,000 x-ray images (radiographs) randomly collected from daily routine work at the Department of Diagnostic Radiology of the RWTH Aachen University\footnote{http://www.rad.rwth-aachen.de/} \cite{Lehmann2003,tommasi2010overview}. All images are classified into 193 categories (classes) and annotated with the IRMA code which relies on class-subclass relations to avoid ambiguities in textual classification \cite{lehmann2003irma,Lehmann2006}. The IRMA code consists of four mono-hierarchical axes with three to four digits each: the technical code T (imaging modality), the directional code D (body orientations), the anatomical code A (the body region), and the biological code B (the biological system examined). The complete IRMA code subsequently exhibits a string of 13 characters, each in $\{0,\dots,9;a,\dots,z\}$:  TTTT-DDD-AAA-BBB. More information on the IRMA database and code can be found in \cite{Lehmann2003,Lehmann2006}. IRMA dataset offers 12,677 images for training and 1,733 images for testing. Figure \ref{fig:IRMASamples} shows some sample images from the dataset along with their IRMA code in the format  TTTT-DDD-AAA-BBB. IRMA x-rays images are a challenging benchmarking case mainly due to the imbalanced class distribution. 

\textbf{Error Measurements --} We used the evaluation scheme provided by ImageCLEFmed09 to compute the difference between the IRMA codes of the testing image and the first hit retrieved by the proposed approach. The total error for all test images can then be calculates as follows \cite{Lehmann2003,Lehmann2006}\footnote{We used the Python implementation provided by ImageCLEFmed09 to compute the errors according to this scheme: http://www.imageclef.org/}:
\begin{equation} \label{equation:IRMA1}
E_\textrm{total} = \sum_{m=1}^{1733} \sum_{j=1}^{4} \sum_{i=1}^{l_{j}} \frac{1} {b_{l_{j},i}} \frac {1} {i} \delta (I_{l_{j},i}^{m}, \tilde{I}_{l_{j},i}^{m})
\end{equation}
Here, $m$ is an indicator to each image, $j$ is an indicator to the structure of an IRMA code, and $l_{j}$ refers to the number of characters in each structure of an IRMA code.
For example, in the IRMA code: 1121-4a0-914-700, $l_{1}=4$, $l_{2}=3$, $l_{3}=3$ and $l_{4}=3$. 
$i$ is an indicator to a character in a particular structure. 
Here, $l_{2,2}$ refers to the character ``a'' and $l_{4,1}$ refers to the character ``7''. 
$b_{l_{j},i}$ refers to the number of branches, i.e. number of possible characters, at the position $i$ in the $l_{j}^{th}$ structure in an IRMA code. 
$I^{m}$ refers to the $m^{th}$ testing image and $\tilde{I}^{m}$ refers to its top 1 retrieved image. 
$\delta (I_{l_{j},i}^{m}, \tilde{I}_{l_{j},i}^{m})$ compares a particular position in the IRMA code of the testing image and the retrieved image. 
It then outputs a value in \{0, 1\} according to the following rules:
\begin{equation} \label{equation:IRMA2}
\delta (I_{l_{j},i}^{m}, \tilde{I}_{l_{j},i}^{m})=
\begin{cases}
0, & I_{l_{j},h}^{m} = \tilde{I}_{l_{j},h}^{m} \forall h \leq i \\
1, & I_{l_{j},h}^{m} \neq \tilde{I}_{l_{j},h}^{m} \exists h \leq i
\end{cases}
\end{equation}

\textbf{Parameters for Autoencoder --} In our experiments, we used the mean squared error as the quadratic cost function. The network was trained using stochastic gradient descent on  training images for 300 epochs using a mini-batch size of 10 images. The learning rate was set to 0.5. No regularization techniques were implemented. Single hidden layer (shallow layer) and the multiple layers (deep layer) are both employed in the experiments. Each hidden layer reduces or increases the number of neurons from the previous layer by a factor of 2. For instance, in the case of the deep autoencoder with 3 hidden layers, it compresses the input by a factor of 4 in the second hidden layer and raises to the half size of input in the third layer (decoder).

\textbf{Performance --} To validate the performance of the proposed method, the Radon barcodes\footnote{Matlab code from http://tizhoosh.uwaterloo.ca/} are generated as well for comparing the error score computation according to Eq. \ref{equation:IRMA1}. A total of 12,677 images in the training dataset are used to generate Radon barcodes (RBC) and the autoencoded Radon barcodes (ARBC). The remaining 1,733 images are used to retrieve the top hit image by generated barcode through $k$-NN direct search ($k$=1) with minimum Hamming distance. 

Table \ref{Table_rbc_arbc} represents the error performance for RBC and ARBC with various normalized image sizes and number of projection angles $n_\theta$. 
From Table \ref{Table_rbc_arbc}, it can be clearly seen that although the length of the ARBC has been compressed, it is able to achieve better results than RBC. And generally, the ARBC compressed to half size works slightly better than the one compressed by a factor of 4. The increase in image size and the number of Radon projections seems to provide better results, but leads to a longer barcode. 

\begin{table}
\centering
\caption{Error for RBC vs. ARBC with one hidden layer of $1/4$ or $1/2$ of size of input layer.}
\begin{tabular}{|l||c|c||c|c|} \hline
Downsampling& \multicolumn{2}{c||}{$32 \times 32$} & \multicolumn{2}{c|}{$64\times 64$} \\ \hline
 $n_\theta$ &8 & 16 & 8 & 16 \\  \hline\hline
RBC & 605.83 & 576.45 & 585.76 & 559.46\\ \hline
ARBC$_{1/2}$ & 474.83 & 465.14 & 493.37 & 426.06\\ \hline
ARBC$_{1/4}$ & 477.27 & 475.14 & 477.32 & 440.69\\ \hline
\hline\end{tabular}
\label{Table_rbc_arbc}
\end{table}

The ARBC generated by multi-layers autoencoders were also implemented. Table \ref{Table_arbc_diff_layer} shows error comparison among several ARBC versions generated from two shallow hidden layers (one in encoding and the other in decoding) and deep hidden layer. Each hidden layer reduces or increases the number of neurons from the previous layer by a factor of 2.

\begin{table*}
\centering
\caption{Error for different layers/settings for Radon transform when three hidden layers are used}
\begin{tabular}{|l|c|c|c|c|} \hline
 & \multicolumn{2}{c|}{$32 \times 32$} & \multicolumn{2}{c|}{$64\times 64$} \\ \hline
Hidden Layer &   8 Projections & 16 Projections & 8 Projections & 16 Projections\\ \hline
1st(encoder) & 519.52 & 453.33 & 466.92 & 392.09 \\ \hline
2nd(deep) & 509.70 & 463.76 & 484.57 & 415.93 \\\hline
3rd(decoder) & 679.54 & 623.16 & 624.23 & 583.21\\ 
\hline\end{tabular}
\label{Table_arbc_diff_layer}
\end{table*}

It can be observed that for the multi-layer autoencoder, the performance is similar to the single layer autoencoder, namely that the error score drops proportional to the increase of the normalized image size and the number of projection angles. It seems that the number of projection angles has a greater impact on error performance than image downsampling size. Obviously, more projection angles bring additional image information. In contrast, the increase in image size (less downsampling) only enhances the information from the original image for certain angles. It can also be observed that the dataset generated by the shallow layer (the first hidden layer) performs slightly better than the deep layer one (the second hidden layer). However, it should be mentioned that the size of the barcode from the deep layer is half the size of the one in the shallow layer, which is $1/4$ the size of the Radon barcode.

\begin{table}[t]
\centering
\caption{Time comparison (in seconds)}
\begin{tabular}{|c|c|c|c|c|} \hline
Downsampling & \multicolumn{2}{c|}{$32 \times 32$ } & \multicolumn{2}{c|}{$64\times 64$ } \\ \hline
$n_\theta$  &8 & 16 & 8 & 16 \\  \hline \hline
%RBC &  0.006 & 0.006 &  0.006 & 0.007 \\ \hline
%ARBC$_{1/2}$ & 0.80 & 2.89 & 2.53 & 18.02\\ \hline
%ARBC$_{1/4}$ & 0.46 & 1.51 & 1.43 & 6.14\\ \hline
%ARBC\_3 & 1.21 & 3.8 & 3.67 & 22.63\\ 
RBC	& 0.006	& 0.006	& 0.006	 & 0.007 \\ \hline
ARBC$_{1/2}$ 	& 0.243	& 0.869	& 0.762	& 5.408 \\ \hline
ARBC$_{1/4}$	& 0.139	& 0.455	& 0.431	& 1.843 \\ \hline
ARBC\_3	& 0.364	& 1.163	& 1.102	& 6.791 \\
\hline\end{tabular}
\label{Table_time}
\end{table}

Table \ref{Table_time} represents the processing time for generating the RBC and the proposed autoencoded Radon barcode (ARBC). The time for ARBC is only shown in one epoch and autoencoder with 3 hidden layers is denoted as ARBC\_3 (1st hidden layer size: $1/2$ of RBC, 2nd :$1/4$ size of RBC, 3rd: $1/2$ size of RBC). From Tables \ref{Table_rbc_arbc} and \ref{Table_arbc_diff_layer}, it can be seen that larger normalized image sizes and more projections do deliver better results in general. However, they also increase the computational costs. We used a  MacBook Pro with a 2.5 GHz quad core intel core i7 (16 GB RAM). We did not use GPUs.  

\textbf{Barcodes versus SURF and BRISK --} In this series of experiments, we also examined SURF \cite{Bay2008} (as a non-binary method) and BRISK \cite{Leutenegger2011} (as a binary method). Using $k$-NN as before was not an option because initial experiments took considerable time as SURF and BRISK appear to be slower than barcodes. Hence, we used locality-sensitive hashing (LSH) \cite{Indyk1998} to hash the features/codes into patches of the search space that may contain similar images\footnote{Matlab code: http://goo.gl/vFYvVJ}. We made several tests in order to find a good configuration for each method. As well, the configuration of LSH (number of tables and key size for encoding) was subject to some trial and errors. We set the number of tables for LSH to 30 (with comparable results for 40) and the key size to a third of the feature vectors' length. We selected the top 10 results of LSH and chose the top hit based on highest correlation with the input image for each method. The results are reported in Table \ref{tab:SURFBRISK}.

As apparent from the results, the main drawback of both SURF and BRISK for medical image retrieval is that they do fail to extract descriptors for some images. BRISK, as a binary descriptor, has very high error rate. Although SURF has lower error than RBC, its non-binary descriptors does constitute a  major disadvantage due to high storage requirement for big image data. ARBC, however, does provide the lowest error rate, and much lower storage needs.  

\begin{table}[]
\centering
\caption{Comparing barcodes with SURF and BRISK. }
\label{tab:SURFBRISK}
\begin{tabular}{l|l|l}
Method & Total Error & Failure \\ \hline
SURF & 526.05 & 4.6\% (79 images) \\
BRISK & 761.96 & 1.1\% (19 images) \\
RBC & 559.46 & 0.0\% \\ 
ARBC & 392.09 & 0.0\% \\ 
\end{tabular}
\end{table}

\section{Conclusions}
Tagging images with binary descriptors seem to be a powerful approach to image retrieval for big image data. Radon barcodes, recently introduced, constitute an interesting framework for generating content-based binary features, probably a suitable approach for tagging medical images. Many aspects of Radon barcodes still require investigations. Among others, the method for binarizing Radon projections has a significant effect on the descriptiveness of the barcode.   
In this paper, we proposed autoencoded Radon barcodes. By making use of an autoencoder with 1 or 3 hidden layers, Radon barcodes can be generated via thresholding of compressed projections accessed at the output of hidden layers. Using IRMA dataset with  14,410 x-rays images, we demonstrated that thresholding via autoencoders is superior to local thresholding via median of non-zero projection values. The experimental results show that the barcodes generated by deeper autoencoder achieve better performance than shallow networks.

\bibliographystyle{abbrv}
\bibliography{sigproc}  % sigproc.bib is the name of the 
\end{document}